\documentclass[11pt,letterpaper]{article}
\usepackage{emnlp2017}
\usepackage{times}
\usepackage{latexsym}

\emnlpfinalcopy

\def\emnlppaperid{***}

\usepackage{qtree}
\usepackage{csquotes}
\usepackage{url}
\usepackage{lingmacros}
\usepackage{booktabs}
\usepackage{graphicx}
\usepackage{subcaption}
\usepackage{bm}
\usepackage{amssymb}
\usepackage{amsmath}
\usepackage{multirow}
\usepackage{pifont}% http://ctan.org/pkg/pifont
\usepackage{footmisc}
\DefineFNsymbols{mySymbols}{{\ensuremath\dagger}{\ensuremath\ddagger}\S\P
   *{**}{\ensuremath{\dagger\dagger}}{\ensuremath{\ddagger\ddagger}}}
\setfnsymbol{mySymbols}
\newcommand{\cmark}{\ding{51}}%
\newcommand{\xmark}{\ding{55}}%

\usepackage{makecell}
\definecolor{darkgreen}{RGB}{51, 153, 51}
\definecolor{darkred}{RGB}{204, 0, 0}
\DeclareMathOperator*{\argmax}{arg\,max}

\title{A Mention-Ranking Model for Abstract Anaphora Resolution 
}

\author{Ana Marasovi\'c, Leo Born\Thanks{Leo Born, Juri Opitz and Anette Frank contributed equally to this work.}, Juri Opitz\footnotemark[1], \and Anette Frank\footnotemark[1]\\
	    Research Training Group AIPHES\\
	    Department of Computational Linguistics\\
	    Heidelberg University\\
	    69120 Heidelberg, Germany\\
	     {\tt \{marasovic,born,opitz,frank\}@cl.uni-heidelberg.de}
} 

\date{}

\begin{document}
	\maketitle
	\begin{abstract}
		Resolving abstract anaphora is an important, but difficult task for text understanding. 
%\rednote{With recent advances in representation learning, it becomes a tangible aim.}
%		. 
		Yet, with recent advances in re\-pre\-sen\-ta\-tion learning this task
		%resolving this type of anaphora 
		becomes a more tangible aim.
		A central property of {\em ab\-stra\-ct anaphora} is that it establishes a relation between the {\em anaphor} embedded in the {\em anaphoric sentence} and its (ty\-pi\-cal\-ly 
		%sentential, or
		non-nominal) {\em antecedent}. 
		We pro\-po\-se %an LSTM\--based 
		a mention-ranking model that learns how abstract anaphors relate to their an\-te\-ce\-dents 
		% was: antecedent sentences
		%that learns the relation between anaphoric and antecedent sentences
		with an LSTM-Siamese Net.
		%network 
		%architecture. 
		We over\-come the lack of training data by ge\-ne\-ra\-ting 
		%large numbers of 
		artificial anaphoric sentence--antecedent pairs. Our model outperforms state-of-the-art results on {\em shell noun re\-so\-lu\-tion}. %dataset. 
		We  also report first benchmark results on an abstract anaphora subset of the ARRAU corpus.
		This corpus presents a greater challenge due to a mixture of nominal and pronominal anaphors and 
		a greater range of confounders. We found model variants that outperform the baselines for nominal anaphors, without training on individual anaphor data, but still lag behind for pronominal anaphors. 
		Our model selects syntactically plausible candidates and -- if disregarding syntax -- discriminates candidates using deeper features. %\bluenote{AF: first: characterize results (over BL for all and nominals only), then add interpretation (above: valid?) and analysis (below)} 
			%Deeper inspection shows that the model is able to learn a relation  between the anaphor in the anaphoric sentence and its antecedent.
%			\\
%			Our results
%			 show that the model learns to identify the true antecedent using sentence representations without the guide of syntactic information.}

	\end{abstract} 
	
	\section{Introduction}

%\rednote{Deeper inspection shows that the model is able to learn a relation  between the anaphor in the anaphoric sentence and its antecedent.}
 
Current research in anaphora (or coreference) resolution is focused on %anaphora 
%\rednote{anaphora -$>$ anaphors; still think "noun phrases" is clear since the anaphors {\bf refer} to noun phrase antecedents (whereas later we talk about nominal/pronominal (abstract) anaphors)}
%(i.e., noun phrases
resolving noun phrases referring to concrete objects or entities in the real world, which is arguably the most %important and most
frequently occurring type.\ 
Distinct from these are diverse types of {\em abstract} anaphora (AA) \cite{Asher:abstractobj} where reference is made to propositions,  facts, events or properties. An example %of abstract anaphora
is given in (1) below.\footnote{Example drawn from ARRAU \citep{uryupina2016arrau}.% ARRAU characterizes (\ex{1}) as an {\em abstract} nominal anaphor and (\ex{2}) as a pronominal anaphor referring to an {\em action} or {\em plan}.
  	}
 
%Other types of anaphora include \textit{propositional} anaphora\footnote{Also referred to as \textit{abstract} or \textit{event} anaphora, or \textit{discourse deixis}. See e.g. \citet{Asher:abstractobj}.}, which refer to an {\em abstract object}, such as a proposition, a property, an event, or a fact. 

%\begin{displayquote}\textit{A very nice example. :)}
%\end{displayquote}

While recent approaches address
%restrict themselves to 
the 
%anaphoric 
resolution of selected abstract {\em shell nouns} \citep{kolhatkar-hirst:2014:EMNLP2014}, 
we aim
%the aim of this paper is 
to resolve a wide range of abstract anaphors, such as the NP
%noun 
\textit{this trend} in (1), 
%and similar cases with 
as well as pronominal anaphors %, such as,
(\textit{this}, \textit{that}, or \textit{it}).%\footnote{We don't capture ellipsis or other special forms.}   

%Due to shortage of
%limitations in %available 
%resources for evaluation, we evauate our models
% on {\em shell noun anaphora} and a selection of {\em abstract anaphora} types drawn from the ARRAU corpus. 
 
%In what follows,
Henceforth, we refer to a sentence that contains an abstract anaphor as {\em the anaphoric sentence (AnaphS)}, and to a constituent that the anaphor refers to as {\em the antecedent (Antec)} (cf.\ (\ex{1})).

\enumsentence{ \small
	Ever-more powerful desktop computers, designed with one or more microprocessors as their "brains", are expected to increasingly take on functions carried out by more expensive minicomputers and mainframes.
	"[$_{Antec}$ {\em \textbf{The guys that make traditional hardware are really being obsoleted by microprocessor-based machines}}]", said Mr.\ Benton. [$_{AnaphS}$ As a result of {\em \textbf{this trend}}$_{AA}$, longtime powerhouses 
	HP, IBM and Digital 
	Equipment Corp.\ 
	are scrambling to  
	counterattack with microprocessor-based systems of their own.]%\rednote{we don't explain AA -$>$ Anaph? $\Rightarrow$ AM: added AA in the previous paragraph, and AntecS later where the anaphoric sentence occurs for the first time. Good? }
		
}

A major obstacle for solving this task is the lack of sufficient amounts of 
annotated training data.
%that no substantial amount of annotated data is available to 
%train sophisticated models. 
%For this reason, 
We propose a method to generate large amounts of training instances covering a wide range of abstract anaphor types. 
%A substantial amount of data 
This 
%will 
enables us to use 
%heavily data-driven, 
neural methods which have shown great success in related tasks:
%such as 
coreference resolution \cite{clark2016deep}, textual entailment \cite{bowman-EtAl:2016:P16-1},
%rocktaschel2015reasoning,
%wang-jiang:2016:N16-1}, 
learning textual similarity \cite{mueller2016siamese}, and discourse relation sense classification %\cite{mihaylovfrank:2016,rutherfordetal:17}.
\cite{rutherfordetal:17}. 

Our model is inspired by the mention-ranking model for coreference resolution \cite{wiseman2015learning, clark2015entity, clark2016deep, clark2016improving} and combines it with a Siamese Net \cite{mueller2016siamese}, %(Mueller and Thyagarajan, 2016)
\cite{ neculoiu-versteegh-rotaru:2016:RepL4NLP} for learning si\-mi\-la\-ri\-ty between sentences. 
%For our task, g
Given an \textit{anaphoric sentence} (AntecS in (1)) 
and a candidate antecedent (any constituent in a given context, e.g.\ \textit{being obsoleted by microprocessor-based machines} in (1)), the LSTM-Siamese Net learns representations for the candidate and the anaphoric sentence in a shared space. 
%\rednote{[AM: added def. of the anaphoric sentence and a brief explanation of a candidate.] \bluenote{AF: inserted separate definition above -$>$ shorten this passage here, see [].}}
%, using information 
%about their being in an anaphoric relation. 
These representations are combined into a joint representation used to calculate a score that characterizes the relation between them. 
The learned score is used to select the highest-scoring antecedent candidate for the given anaphoric sentence and hence its anaphor. 
%their representations using information 
%about their being in an anaphoric relation. 
%We consider one anaphor at a time and mark it using its surrounding context (cf.\ \cite{zhou-xu:2015:ACL-IJCNLP}).%Currently we assume a 1:1 correspondence between the anaphoric sentence and
We consider one anaphor at a time and provide the embedding of the {\em context of the anaphor} and the embedding of the {\em head of the anaphoric phrase} to the input to characterize each individual anaphor -- similar to the encoding proposed by \citet{zhou-xu:2015:ACL-IJCNLP} for individuating multiply occurring predicates in SRL. With deeper inspection we show that the model learns a relation between the anaphor in the anaphoric sentence and its antecedent. %Currently we assume a 1:1 correspondence between the anaphoric sentence and
%\rednote{[AM: different wording of the last sentence, required reading the full ZHOU\&XU paper to understand it.] \bluenote{AF: see above}} 
Fig.\ \ref{fig:architecture} displays our architecture.

In contrast to other work, 
our method for generating training data is not confined to specific types of anaphora such as shell nouns \citep{kolhatkar-hirst:2014:EMNLP2014} or anaphoric connectives \citep{stede:grishina:2016:demzufolge}. It produces large amounts of instances and is easily adaptable to other languages. This enables us to build a robust, knowledge-lean model for abstract anaphora resolution that
% which uses a minimal amount of external resources, and for this reason, 
easily extends to multiple languages. 

We evaluate our model on the shell noun resolution dataset of \citet{kolhatkar-zinsmeister-hirst:2013:EMNLP}
%\footnote{We thank the authors for making this data available to us.}
% \citep{kolhatkar-hirst:2014:EMNLP2014} 
and show that it outperforms their state-of-the-art results. 
%\rednote{TBD: We also evaluate on their data using (in addition) our artificially constructed training resources?}
Moreover, we report results of the model (trained on our newly constructed dataset) on unrestricted abstract anaphora instances from the
% manually annotated 
ARRAU corpus \cite{POESIO08.297,uryupina2016arrau}. To our knowledge this provides the first state-of-the-art benchmark on this data subset.

Our TensorFlow\footnote{\citet{tensorflow2015-whitepaper}} implementation of the model and scripts for data extraction are available at: \url{https://github.com/amarasovic/neural-abstract-anaphora}.

	\section{Related and prior work} 

{\bf Abstract anaphora} has been extensively studied in linguistics and shown to exhibit specific properties in terms of semantic antecedent types, their degrees of abstractness, and general discourse properties \citep{Asher:abstractobj,webber1991structure}. In contrast to nominal anaphora, abstract anaphora is difficult to resolve, given that agreement and lexical match features are not applicable. Annotation of abstract anaphora is also difficult for humans \citep{DipperZinsmeister:12_11}, and thus, only few smaller-scale corpora have been constructed. We evaluate our models on a subset of the ARRAU corpus \citep{uryupina2016arrau} that contains abstract anaphors and the shell noun corpus used in \citet{kolhatkar-zinsmeister-hirst:2013:EMNLP}.\footnote{We thank the authors for making their data available.} 
We are not aware of other freely available abstract anaphora datasets.

Little work exists for the {\bf automatic resolution of abstract anaphora}. 
Early work \citep{EckertStrube:00,strube-muller:2003:ACL,Byron:04,mueller:08} has focused on spoken language, which exhibits specific properties.
%\footnote{The datasets are not freely available.} 
Recently, {\bf event coreference} has been addressed  using feature-based classifiers \cite{jauhar-EtAl:2015:*SEM2015,LuNg:16}. %[However, event coreference
%phenomena 
%differs in important properties from %typical cases of 
%abstract anaphora]. % resolution.
%, which is the focus of our work. 
Event coreference is restricted to a subclass of {\em events}, and usually focuses on coreference between verb (phrase) and noun (phrase) mentions of similar abstractness levels (e.g.\ {\em purchase -- acquire}) with no special focus on (pro)nominal anaphora. Abstract anaphora typically involves a full-fledged clausal antecedent that is referred to by a highly abstract (pro)nominal anaphor, as in (\ex{0}).

\citet{rajagopal2016unsupervised} proposed a model for resolution of events in biomedical text that refer to a single or multiple clauses.
However, instead of selecting the correct antecedent clause(s) (our task) for a given event, their model is restricted to classifying the event into six abstract categories: this these \textit{changes, responses, analysis, context, finding, observation}, based on its surrounding context. While related, their task is not comparable to the full-fledged abstract anaphora resolution task, since the events to be classified are known to be coreferent and chosen from a set of restricted abstract types.

More related to our work is \citet{anand-hardt:2016:EMNLP2016} who present an antecedent ranking account for {\bf sluicing} using classical machine learning based on a small training dataset. They employ features modeling distance, containment, discourse structure, and -- less effectively -- content and lexical correlates.\footnote{Their data set was not publicized.} 

Closest to our work is %the work of 
\citet{kolhatkar-zinsmeister-hirst:2013:EMNLP} (KZH13) and \citet{kolhatkar-hirst:2014:EMNLP2014} (KH14) on \textbf{shell noun resolution}, using classical machine learning techniques.
% (henthforth referred to as {\bf KZH13} and {\bf KZH14}). 
%{\em Shell noun anaphora} considers a restricted form of anaphors, such as {\em the fact/issue that/whether}, which encapsulate a specific type of abstract  meaning, indicated by the noun. 
%Among machine-learning based resolution systems for \textit{non-nominal} antecedents the most prominent is resolution of \textit{shell nouns}.
Shell nouns are abstract nouns, such as \textit{fact}, \textit{possibility}, or \textit{issue}, which can only be interpreted 
jointly 
%together 
with their \textit{shell} content (their embedded clause as in (\ex{1}) or antecedent as in (\ex{2})). 
% i.e.,\ the embedded sentence content they encapsulate in the given context, 
%The same shell nouns 
%Shell nouns 
%may be {\em anaphoric} to an abstract object in the prior discourse, as in (\ex{2}).
KZH13 refer to shell nouns whose antecedent occurs in the prior discourse as \textit{anaphoric shell nouns} (ASNs) (cf.\ (\ex{2})), and {\em cataphoric shell nouns} (CSNs) otherwise (cf.\ (\ex{1})).\footnote{We follow this terminology for their approach and data representation.} %\rednote{[AM: added a -simple- explanation of ASN/CSN.]} \bluenote{fine}
%or  \textit{cataphoric} (\ex{2}).

%\enumsentence{\small
%%	Cataphoric shell noun (CSN):
% Congress has focused almost solely
% on
%		{\bf\em \underline{the fact}}
%			that {\em special education is expensive - and that
%			it takes away money from regular education.}}
%
%%\vspace*{-7mm}
%\enumsentence{\small
%%Anaphoric shell noun (ASN): 
%Environmental
%	Defense [..]  notes
%	that {\em “Mowing the lawn with a gas mower produces
%	as much pollution in half an hour as driving a
%	car 172 miles.”}   {\bf\em \underline{This fact}}
%	may [..]
%	%help to 
%	explain the  recent  surge  in  the  sales  of  [..] 
%	%the  good  
%	old-fashioned  push  mowers  or  the  battery-powered
%	mowers.}
	
\enumsentence{\small
    %    Cataphoric shell noun (CSN):
    Congress has focused almost solely
    on
    {\bf\em the fact}
    that [{\em special education is expensive - and that
        it takes away money from regular education.}]}
\enumsentence{\small
    %Anaphoric shell noun (ASN):
    Environmental
    Defense [...]  notes
    that [$_{Antec}$ {\em “Mowing the lawn with a gas mower produces
        as much pollution [...] as driving a
        car 172 miles.“}]   [$_{AnaphS}$ {\bf\em This fact}
    may [...]
    %help to
    explain the  recent  surge  in  the  sales  of  [...]
    %the  good  
    old-fashioned  push  mowers [...]].}
    % or  the  battery-powered
    %mowers.]}

%\citet{kolhatkar-zinsmeister-hirst:2013:EMNLP} 
KZH13 presented an approach for resolving six typical shell nouns following the observation that 
%cataphoric 
%shell noun occurrences as in (\ex{-1})
CSNs are easy to resolve based on their syntactic structure alone, and the assumption that ASNs share linguistic properties with their embedded (CSN) counterparts.
%cataphoric and 
%anaphoric antecedents as in (\ex{0})
%CSNs and ASNs share linguistic properties with their embedded counterparts. \bluenote{AF:should this rather read as: "that ASNs share linguistic properties with their embedded (CSN) counterparts"? [or are they also "resolve" CSNs?]} 
%\footnote{They call constructions (\ex{-1}) {\em cataphoric shell nouns (CSN)} and those in (\ex{0}) {\em anaphoric shell nouns (ASN)}.}
%\rednote{[AM: SVM-rank is trained to resolve CSNs and then this model is used to resolve ASNs.]} 
They manually developed rules to identify the embedded clause (i.e.\ cataphoric antecedent) of %cataphoric shell nouns (cf.\ (\ex{-1}))
 CSNs and trained SVM$^{rank}$ \cite{joachims2002optimizing} on such instances.
 %. Later, they used the trained candidate-ranking model 
 %to resolve 
 %%more complex
%%  anaphoric shell nouns (cf.\ (\ex{0}))
  %ASNs
  %Finally, t
%  \rednote{old: finally,} 
  The trained SVM$^{rank}$ model is then used to resolve ASNs.  %\footnote{We follow KZH13 and KZH14 who established this terminology for their approach and data representation.}
  KH14 generalized their method to be able 
  % replace the manually developed rules with a general method %that can be used 
   to create training data for any given shell noun, 
   %and consequently to resolve harder anaphoric cases. 
   however, their method heavily exploits the specific properties of shell nouns and does not apply to other types of abstract anaphora. 
   %and their categorization.
   % described in the linguistics literature. 
%\rednote{[AM: used abbreviations for CSNs/ASNs.]} \bluenote{fine}
%A related phenomenon that could be deployed for accessing antecedents of abstract anaphors is explored in \citet{stede:grishina:2016:demzufolge} for German.

\citet{stede:grishina:2016:demzufolge} study a related phenomenon for German. They examine inherently anaphoric connectives 
%that are inherently anaphoric 
(such as {\em demzufolge -- according to which}) that
%, and thus 
could be used to access their abstract antecedent in the immediate context. Yet, such connectives are restricted in type, and the study shows that such connectives are often ambiguous with nominal anaphors and 
%also
require sense disambiguation. We conclude that they cannot be easily used to acquire antecedents automatically.

In our work, we explore a different direction: we construct artificial training data using a general pattern that identifies embedded sentence constituents, which allows us to extract relatively secure training data for abstract anaphora that captures a wide range of anaphora-antecedent relations, and apply this data to train a model for the resolution of unconstrained abstract anaphora.

Recent work in entity coreference resolution has proposed powerful neural network-based models that we will adapt to the task of abstract anaphora resolution. Most relevant for our task is the \textbf{mention-ranking neural coreference model} proposed in \citet{clark2015entity}, and their improved model in \citet{clark2016deep}, which integrates a loss function \citep{wiseman2015learning} which learns distinct feature representations for anaphoricity detection and antecedent ranking.
%\citep{clark2016improving}-other

\textbf{Siamese Nets} %is an architecture for metric learning. %, i.e.\ i
%It 
%learn to 
distinguish between similar and dissimilar pairs of samples by optimizing a loss %function
over the metric induced by the representations. %In classical classification tasks, underlying matrices are treated as side effects and are not explicitly sought.
It is widely used in vision %utilized 
%with
%with ConvNets
 \cite{chopra2005learning}, %Recently, i
%It has been used 
and 
%recently 
in NLP for semantic similarity, entailment, query normalization and QA %question answering 
\citep{mueller2016siamese, neculoiu-versteegh-rotaru:2016:RepL4NLP, Das2016TogetherWS}. 
%Rest:
%[\textbf{Annotation of PA} Strube and Mueller (2003);
%Botley (2006) \citep{Botley:06} ] 

	\section{Mention-Ranking Model}
\begin{figure}
\begin{center}
\includegraphics[height=0.32\textheight]{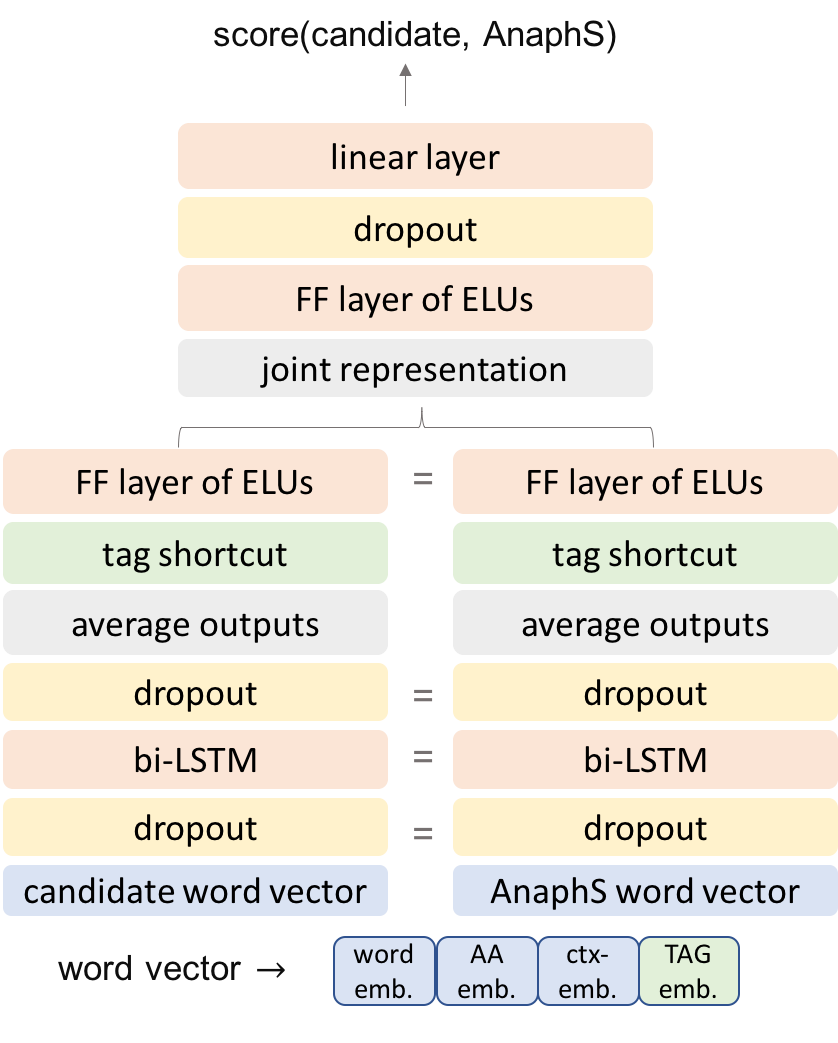}
\caption{Mention-ranking architecture for abstract anaphora resolution (MR-LSTM).}
\vspace*{-3mm}
\label{fig:architecture}
\end{center}
\end{figure}

%\rednote{Give a better high-level explanation of what does the model do.} 
Given an anaphoric sentence $s$ with a marked anaphor (mention) 
and a candidate antecedent $c$, the mention-ranking (MR) model assigns  the pair $(c,s)$ a score, using re\-pre\-sen\-ta\-tions produced by an LSTM-Siamese Net.
%neural network. 
The highest-scoring candidate is as\-sig\-ned to the marked anaphor in the anaphoric sentence. %
%the anaphoric sentence with its marked
%%, and hence its 
%anaphor. \bluenote{better: to the marked anaphor in the anaphoric sentence.} 
Fig.\ \ref{fig:architecture} displays the model.
%The proposed architecture is outlined in 

We learn representations of an anaphoric sentence $s$ and a candidate antecedent $c$ using a %variant of the 
bi-directional
Long Short-Term Memory \cite{hochreiter1997long, Graves2005FramewisePC}.
%, known as a bi-directional LSTMs.
%It produces two representations of a sentence, by processing it from left to right and from right to left. The obtained representations are combined, most frequently, by taking a linear combination or concatenation. 
%In the architecture we propose,
One bi-LSTM is applied to the anaphoric sentence $s$ and a candidate antecedent $c$, hence the term siamese. %\rednote{shortened (see comments)}
Each word is represented with a vector $\boldsymbol{w_i}$
%, which is 
constructed by concatenating embeddings of
%an embedding of 
the word, 
of the context of the anaphor (average of embeddings of the anaphoric phrase, the previous and the next word),
of the head of the anaphoric phrase\footnote{Henceforth we refer to it as embedding of the anaphor.}, and, finally,
%an embedding 
an embedding of the constituent tag of the candidate, or 
%an embedding of 
the \textit{S} constituent tag if the word is in the anaphoric sentence.
For each sequence $s$ or $c$,
%proposition, 
the %corresponding 
word vectors $\boldsymbol{w_i}$ are sequentially fed into the bi-LSTM, which produces outputs
%, i.e., cells after the output gate, 
from the forward pass, $\overrightarrow{\boldsymbol{h_i}}$, and outputs $\boldsymbol{\overleftarrow{h_i}}$ from the backward pass. The final output of the i-th word is defined as $\boldsymbol{h_i} = [\boldsymbol{\overleftarrow{h_i}}; \boldsymbol{\overrightarrow{h_i}}]$. To get a representation of 
%the proposition, $\boldsymbol{h_p}$,  
the full sequence, $\boldsymbol{h_s}$ or $\boldsymbol{h_c}$,  
all outputs are averaged, except for those that
%the outputs that 
correspond to padding tokens. 

To prevent forgetting 
%highlight
%assure that 
the constituent tag of the sequence,
% is not forgotten, 
we concatenate %the constituent tag embedding of the sequence
the corresponding tag embedding with $\boldsymbol{h_s}$ or $\boldsymbol{h_c}$ (we call this a
%We refer to this as a 
\textit{shortcut} for the tag information). The resulting vector is fed into a feed-forward layer of exponential linear units (ELUs) \cite{Clevert2015FastAA} to produce the final representation $\boldsymbol{\tilde{h}_s}$ or $\boldsymbol{\tilde{h}_c}$ of the sequence. % is produced. 

%Let $\boldsymbol{\tilde{h}_c}$ and $\boldsymbol{\tilde{h}_s}$ be representations of a candidate and the anaphoric sentence obtained with the model described so far. 
From %the representations 
$\boldsymbol{\tilde{h}_c}$ and $\boldsymbol{\tilde{h}_s}$ we compute a 
%feature 
vector %$\boldsymbol{h_{c,s}}$ that represents their relatedness as
% $\boldsymbol{\tilde{h}_c}$ and $\boldsymbol{\tilde{h}_s}$ is computed as
$\boldsymbol{h_{c,s}} = [|\boldsymbol{\tilde{h}_c} - \boldsymbol{\tilde{h}_s}|; \boldsymbol{\tilde{h}_c} \odot \boldsymbol{\tilde{h}_s}]$ \cite{Tai2015ImprovedSR}, where  $|\text{--}|$ denotes the absolute values of the element-wise subtraction, and  $\odot$ the element-wise multiplication. Then $\boldsymbol{h_{c,s}}$ is fed into a feed-forward layer of ELUs to obtain the final joint representation, $\boldsymbol{\tilde{h}_{c,s}}$, of the pair $(c,s)$.
Finally, we compute the score for the pair $(c, s)$ that represents relatedness between them, by applying a single fully connected linear layer %of size one
to the joint representation:
\begin{equation}\label{eq:score}
score(c,s) = W\boldsymbol{\tilde{h}_{c,s}} + b \in \mathbb{R},
\end{equation}
where W is a $1 \times d$ weight matrix, and $d$ the dimension of the vector $\boldsymbol{\tilde{h}_{c,s}}$.

We train the described mention-ranking model with the 
%version of the 
max-margin training objective from \citet{wiseman2015learning}, used for the antecedent ranking subtask. 
%More precisely, 
%suppose that the training set consists of $n$ anaphoric sentences $s_1,\hdots,s_n$.
Suppose that the training set $\mathcal{D} = \{(a_i, s_i, \mathcal{T}(a_{i}), \mathcal{N}(a_{i})\}_{i=1}^{n}$, where $a_i$ is the i-th abstract anaphor, $s_i$ the corresponding anaphoric sentence, $\mathcal{T}$$(a_{i})$ the set of antecedents of $a_{i}$ and $\mathcal{N}$$(a_{i})$ the set of candidates that are not antecedents (negative candidates). Let $\tilde{t_i} = \argmax_{t \in \mathcal{T} (a_{i})} score(t_i, s_i)$ be the highest scoring antecedent of $a_i$. Then the loss is given by
\begin{equation*}
\sum_{i=1}^{n} \max(0,  \max_{c \in  \mathcal{N} (a_{i})} \{1 + score(c, s_i) - score(\tilde{t_i}, s_i)\}).
\end{equation*}

%Let $\mathcal{C}$$(a_{i})$ denote the set of candidates for the i-th anaphor occurring in the anaphoric sentence $s_i,$ and $\mathcal{T}$$(a_{i})$ the subset of its true antecedents.
%Let $\mathcal{T}$$(a_{i})$ denote the set of antecedents of $a_{i}$ and $\mathcal{N}$$(a_{i})$ the set of candidates that are not antecedents (negative candidates).  
	
	%\Tree[.IP [.NP [.Det \textit{the} ]
%               [.N\1 [.N \textit{package} ]]]
%          [.I\1 [.I \textsc{3sg.Pres} ]
%                [.VP [.V\1 [.V \textit{is} ]
%                           [.AP [.Deg \textit{really} ]
%                                [.A\1 [.A \textit{simple} ]
%                                  \qroof{\textit{to use}}.CP ]]]]]]

\section{Training data construction}

\begin{figure}[t]
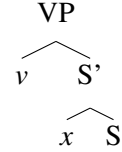

%\Tree[.VP [\textit{v} ]
%         [.S' [\textit{x} ]
%               [.S
%               ]]]                      
\Tree[.VP \textit{v} 
          [.S' \textit{x} 
               [.S
                           ]]]
\caption{A general pattern for artificially creating anaphoric sentence--antecedent pairs. 
	%S is cut off and represents the artificial antecedent $c$. $x$ is replaced with a suitable anaphor, if not already given, to obtain the remnant anaphoric sentence $s$.
	}
\label{fig:pattern}
\end{figure}

We create large-scale  training data for abstract anaphora resolution by exploiting a common construction, consisting of a verb with an embedded sentence (complement or adverbial) (cf.\ 
Fig.\ \ref{fig:pattern}).
%Figure \ref{fig:pattern} displays this very common syntactic pattern, which we exploit to artificially create large-scale training data for abstract anaphora resolution. 
We detect this pattern in a parsed corpus, 'cut off' the S$^{\prime}$ constituent and replace it with a suitable anaphor to create the anaphoric sentence (AnaphS), while S yields
%represents 
the antecedent (Antec).  
%We found that 
This method covers a wide range of 
%potential 
anaphora-antecedent constellations, due to diverse semantic or discourse relations that hold between the clause hosting the verb and the embedded sentence. 

First, the pattern applies to verbs that embed sentential arguments. In
%For example, i
(\ex{1}),  the verb 
{\em doubt} establishes a specific semantic relation between the embedding sentence and its sentential complement. 

%\enumsentence{\textit{The Treasury Department said  [$_{S^{\prime}}$ \textbf{that} [$_S$ the U.S. trade deficit may worsen next year, after two years of significant improvement}]].}
\enumsentence{\small He doubts [$_{S^{\prime}}$[$_S$ a Bismarckian super state will emerge that would dominate Europe], but warns of "a risk of profound change in the [..]
		%heart of the 
		European Community from a Germany that is too strong, even if democratic''].}

From this we extract the artificial antecedent \textit{A Bismarckian super state will emerge that would dominate Europe}, and its corresponding anaphoric sentence \textit{He doubts \textbf{this}, but warns of "a risk of profound change ... even if democratic''}, %, containing the anaphor \textbf{this} (or alternatively \textbf{that}). 
which we construct by randomly choosing one of a pre-defined set of appropriate anaphors (here: {\em this, that, it}), cf.\ Table \ref{tab:sbarhead}. 
%for which the anaphor was randomly chosen between \textit{this}, \textit{that}, \textit{this (that) is true}.
%\rednote{AM: added explanation of the choice of anaphor.}
%\bluenote{AF: better: which we construct by randomly choosing one of a pre-defined set of appropriate anaphors (here: {\em this, that, it}), cf.\ Table \ref{tab:sbarhead}.  [AF: can't make sense of "this (that) is true"]} 
The second row in Table \ref{tab:sbarhead} is used when the head of S$^{\prime}$ is filled by an overt complementizer (\textit{doubts that}), as opposed to (\ex{0}). The remaining rows in Table \ref{tab:sbarhead} apply to adverbial clauses of different types. 

%\rednote{[AM: is the paragraph "Frequent types of sentential adjuncts..." now necessary? - AF: can be omitted; DONE]}
%For example, for: 
%\enumsentence{\small He doubts [$_{S^{\prime}}$ %\textbf{that} [$_S$ a Bismarckian super state will emerge %that would dominate Europe], but warns of "a risk of %profound change in the [..]
%		%heart of the 
%		European Community from a Germany that is too %strong, even if democratic''].}
		%\bluenote{AF: clarify difference between empty and general in Table 1 (?)} \rednote{AM: Done.}
Adverbial clauses encode specific discourse relations with their embedding sentences, often indicated by their conjunctions.
%There are also cases, where the embedded sentence requires a more specific type of abstract anaphora. 
In (\ex{1}), for example, the causal conjunction \textbf{as} relates a cause (embedded sentence) and its effect (embedding sentence):
%indicates that the embedded clause stands in a causal relation with an effect, given by the embedding sentence:
%The first part describes an effect, followed by it's cause. 
%S$^{\prime}$-head  suggests that the following sentence is of causal type:

\enumsentence{\small
		There is speculation that property casualty firms will sell even more munis [$_{S^{\prime}}$ \textbf{as} [$_S$ they scramble to raise cash to pay claims related to Hurricane Hugo [..]
		%and the Northern California earthquake
		]].}

%\begin{displayquote}\textit{There is speculation that property casualty firms will sell even more munis $[$S$^{\prime}$ \textbf{as} $[$S they scramble to raise cash to pay claims related to Hurricane Hugo and the Northern California earthquake$]]$.}
%\end{displayquote}

\noindent
We randomly replace causal conjunctions {\em because, as} 
%In such cases, where the S$^{\prime}$-head is in 
%($\lbrace because,~as \rbrace$), 
with %the chosen, 
appropriately adjusted anaphors, e.g.\ 
\textit{because of that}, \textit{due to this} or \textit{therefore} that make the %respective 
causal relation %(here causal)
 explicit in the anaphor.\footnote{%Type can be ambiguous. While in many cases \textit{as} marks a causal subordinate, it can also %%cases, where 
 	%%its nature can be seen as rather temporal.
 %be interpreted as being temporal. We generally choose the most frequent interpretation. 
 In case of ambiguous conjunctions (e.g. {\em as} interpreted as causal or temporal), we generally choose the most frequent interpretation.}

\begin{table}[t]

\resizebox{0.48\textwidth}{!}{
	\begin{tabular}{@{}lll@{}}
		\toprule
		type & head of S$^{'}$ & possible anaphoric phrase\\
		\midrule
		empty & $\emptyset$ & this, that%\greennote{, this/that is true} 
		\\
		general & that, this & that, this \\
		causal & because, as & therefore, because of this/that,\\
		%&& \greennote{ due to this/that}\\
		temporal& while, since, etc. & during this/that\\
		conditional& if, whether & if this/that is true\\
		\bottomrule
	\end{tabular}
}
	\caption{S$^{'}$-heads and the anaphoric types and phrases they induce (most frequent interpretation).}
	% (for ambiguous conjunctions we choose the most frequent interpretation).}
	\label{tab:sbarhead}
\end{table} 

%For example, from

%\begin{displayquote}\textit{They graze at the Farmers Market, a combination gourmet food court and grocery store, $[$S$^{\prime}$ \textbf{while} $[$S a pianist accompanies the noon fashion show with a selection of dreamy melodies$]]$}.
%\end{displayquote} 

%we construct the source \textit{They graze at the Farmers Market, a combination gourmet food court and grocery store during \textbf{this}.} And the antecedent becomes \textit{A pianist accompanies the noon fashion show with a selection of dreamy melodies.}

Compared to the shell noun corpus of KZH13, who made use of a carefully constructed set of extraction patterns, a downside of our method is that our artificially created antecedents are uniformly of type S. However, the majority of abstract anaphora antecedents found in the existing datasets are of type S. %(or combinations of NP and PP constituents). 
Also, our models are intended to induce semantic representations, and so we expect syntactic form to be less critical, compared to a feature-based model.\footnote{This also alleviates problems with languages
%	\greennote{when applying our method to a language} 
	like German, where (non-)embedded sentences differ in surface position of the finite verb. We can either adapt the order or ignore it, when producing anaphoric sentence -- antecedent pairs.}
%this weights not so heavily, when one considers that most of abstract anaphoras in natural text point to propositions (of type S). 
Finally,
% On the other hand, 
the general extraction pattern in Fig.\ \ref{fig:pattern},
%as opposed to patterns for specific shell nouns,
covers a much wider range of anaphoric types.

Using this method we generated a dataset of artificial anaphoric sentence--antecedent pairs from the WSJ  part of the PTB Corpus \citep{marcus1993building},
automatically parsed using the Stanford Parser \cite{klein2003accurate}. %During the construction process, we verified good quality of the generated data. \greennote{Readers may inspect a randomly chosen} \bluenote{A random} sample set \bluenote{verified to be XX\% correct is given} in the Supplementary Materials.\rednote{[AM:add in the supplement random instances we have checked before]}
%For statistics cf.\ Table \ref{tab:data_statistics}.
 
%a clear advantage is that we can approach the resolution of abstract anaphora's in a more general way: our pattern captures more types of abstract anaphora constellations (e.g.\ temporal relationships) as it doesn't rely on specific \textit{shell-noun + this/that} patterns, but only on the very common syntactic pattern displayed in Figure \ref{fig:pattern}. 
%For each pair ($c,a$)
%%()source, antecedent) pair 
%we create a list of candidates, by extracting all syntactic constituents from a window covering the sentence containing the anaphor and the three preceding sentences. All constituents that are not true antecedents are considered as negative candidates.
%For our experiments, w

%We extract the artificial dataset from the WSJ part of the PTB Corpus \citep{marcus1993building}, automatically parsed using Stanford Parser \cite{klein2003accurate}.

%of different quality and size: (i) one from the gold syntactic annotation of the WSJ  part of the PTB Corpus \citep{marcus1993building}; (ii) another from the same corpus but automatically parsed using Stanford Parser \cite{klein2003accurate};  (iii) we create a larger dataset from automatic parses of the NYT Corpus \citep{sandhaus2008new} (years 2000 -- 2010), which we plan to use in future work. In our experiments we always use (ii).
%: the extractions from the automatically parsed PTB corpus.

	\section{Experimental setup}

%\rednote{[AM: true antecedent, %positives $\Rightarrow$ %antecedents; forced using plural %when possible (ASN, ARRAU)]}
\subsection{Datasets}
%\rednote{Be sure that I mention all preprocessing steps!}

\noindent
We evaluate our model on two types of anaphora: (a) {\em shell noun anaphora} and (b) (pro)nominal abstract 
%{\em abstract} and {\em plan} 
anaphors extracted from ARRAU.
% \rednote{$\Rightarrow$ AM: later "We extracted all abstract anaphoric instances from 462 the WSJ part of ARRAU that are marked with the 463 category abstract or plan" + the footnote \# 14, redundant?}. Statistics for both test sets is given in Table \ref{tab:data_statistics}.

%\begin{table}[t]
%	%\begin{tabular}{@{}lccc@{\hspace*{-1mm}}c@{}}
%	\resizebox{0.48\textwidth}{!}{
%	\begin{tabular}{c|cccc}
%	\toprule
%		
%	\multicolumn{1}{c}{}& size & \multicolumn{2}{c}{median length} & avg.\ dist.\\
%	\multicolumn{1}{c}{}&        &  AnaphS &  AnteS &  \\
%	\midrule
%	ASN  & 2323& 23  & 14 & 1\\
%	ARRAU-P & 600 & 28 & 21 & 2\\
%	\midrule\midrule
%	CSN & 114,753 & 11.5 & 14& n.a. \\
%	Artificial& 15,282& 14& 13& n.a.\\
%	\bottomrule
%	\end{tabular}
%	}
%	\caption{In the upper part is data statistics for the test, and in the lower part for the training datasets. For the ASN and the CSN reported statistics are calculated over all shell nouns, but classifiers are trained independently.}\label{tab:data_statistics}
%\end{table}

\paragraph{a. Shell noun resolution dataset.}\hspace*{-3mm}
For com\-pa\-ra\-bi\-li\-ty we train and evaluate our model for
%on the task of 
{\em shell noun resolution}, using the 
original training (CSN) and test (ASN) corpus of \citet{kolhatkar-zinsmeister-hirst:2013:LAW7-ID, kolhatkar-zinsmeister-hirst:2013:EMNLP}.\footnote{We thank the authors for providing the available data.} \hspace*{-0mm} We follow the data preparation and evaluation protocol of  \citet{kolhatkar-zinsmeister-hirst:2013:EMNLP}
%, henceforth referred to as 
({\bf KZH13}). 
%\bluenote{For strict comparison to KZH13's results, we employ only their original training data.}  

The {\bf CSN corpus} was constructed from the NYT corpus using manually developed patterns to identify the antecedent of cataphoric shell nouns (CSNs). %\footnote{\greennote{Again, we are following KZH13's terminology.}}
In KZH13, all syntactic constituents of the sentence that contains both the CSN and its antecedent were considered as candidates for training a ranking model. %Candidates that differ from the true antecedent in only one word were considered as \textit{positive}\footnote{We obtained this information from the authors directly.}, all other candidates as \textit{negative} samples. 
Candidates that differ from the antecedent in only one word or one word and punctuation were as well considered as antecedents\footnote{We obtained this information from the authors directly.}. To all other candidates we refer to as \textit{negative} candidates. %KZH13 used the full CSN to train  SVM$^{rank}$.
For every shell noun, KZH13 used the corresponding part of the CSN data to train SVM$^{rank}$.
%\footnote{The CSN corpus we obtained from the authors contained tokenized sentences for antecedents and anaphoric sentences. The number of instances differed from the reported numbers in KZH13 in $9$ to $809$ instances for training, and $1$ for testing (cf. Table \ref{tab:results_asn}). %\citet{kolhatkar-zinsmeister-hirst:2013:EMNLP}.
%%Additionally
%The given 
%%anaphoric 
%sentences still contained the true antecedent, so we removed it from the sentence and transformed the corresponding shell %noun was transformed 
%into "\textit{this} $\langle \textit{shell noun} \rangle$". 
%%For example, following transformation was done: 
%An example of this process is:
%\textit{The decision \textbf{to disconnect the ventilator} came after doctors found no brain activity.}\ $\rightarrow$ \textit{This decision came after doctors found no brain activity.}}
%Following KZH13, %\citet{kolhatkar-zinsmeister-hirst:2013:EMNLP} 
%we train and evaluate every shell noun independently.

The {\bf ASN corpus} serves as the test corpus. It was also constructed from
 the NYT corpus, by selecting anaphoric instances with the pattern "\textit{this} $\langle \textit{shell noun} \rangle$" for all covered shell nouns. For validation, \citet{kolhatkar-zinsmeister-hirst:2013:LAW7-ID} crowdsourced annotations for the sentence which contains the antecedent, which KZH13 refer to as 
 a \textit{broad region}. Candidates for the antecedent were obtained by using all syntactic constituents of the broad region as candidates and ranking them using the SVM$^{rank}$ model trained on the CSN corpus. The top 10 ranked candidates were presented to the crowd workers and they chose the best answer that represents the ASN antecedent. The workers were encouraged to select \textit{None} when they did not agree with any of the displayed answers and could provide information about how satisfied they were with the displayed candidates. We consider this dataset as gold, as do KZH13, although it may be biased towards the offered candidates.\footnote{The authors provided us with the workers' annotations of the broad region, antecedents chosen by the workers %evaluations of the ranked candidates
and links to the NYT corpus. The extraction of the anaphoric sentence and the candidates had to be redone.}

\paragraph{b. Abstract anaphora resolution data set.} 
 We use the automatically constructed data 
 %obtained 
 from the WSJ corpus (Section 4) for training.\footnote{We excluded any documents that are part of ARRAU.} 
%using the Stanford Parser, 
%We use our artificial data for training, 
%Additionally, we experiment with extending the set of candidates with candidates obtained from an extended window of preceding sentences.
%We optionally extend it with (a portion of) the ASN corpus. We use the following train/dev setups: (1) our artificial data for training, the whole ASN as dev set, (2) our artificial data enriched with 75\% of the ASN data for training, the remaining 25\% of the ASN as development set.  
%Additionally, we experiment with the following options: (i) pruning negative candidates w.r.t. their constituent tags, and (ii) extending the set of candidates with candidates obtained from an extended window of preceding sentences. For (i) a negative candidate is omitted if its constituent tag is not  $\in$ \{S, SBAR, SBARQ, VP, ROOT, None, UNK\}. 
Our test data for unrestricted abstract anaphora resolution is obtained from the {\bf ARRAU corpus} \citep{uryupina2016arrau}. We extracted all abstract anaphoric instances from the WSJ part of ARRAU that are marked with the category \textit{abstract} or \textit{plan},\footnote{ARRAU distinguishes  {\em abstract} anaphors and (mostly) pronominal anaphors referring to an action or plan, as {\em plan}.} and call the
% resulting
 subcorpus {\bf ARRAU-AA}. %As for ASN, by design, annotation of more than one antecedent for one anaphor is possible in ARRAU-AA.
% \bluenote{AF: Re. Redundancy: I suggest: omit this passage here, keep the more general one in Data statistics below. For ASN it is mentioned in fn 12. It's easily overlooked there, so better take it out there and just state it below in statistics.} 
  %Statistics 
 %%for this data 
 %is given in Table \ref{tab:data_statistics}. 
  
%\paragraph{c. Pre-processing.}
%To use pre-trained word embeddings we had to uncase all the data. As we use an automatic parse to extract all syntactic constituents, due to the parser errors, candidates with a same string appeared with different tags. We eliminated duplicates by checking what tag is more frequent for candidates which have the same POS tag of the first word as the duplicated candidate, in the whole dataset. If duplicated candidates are still occurring, we chose any of them. If such duplicates occur in positive candidates, we don't take such instances in the training data to eliminate noise, or choose any of them for the test data. For the training data we choose instances that have at least $10$ tokens long anaphoric sentence. 

\paragraph{Candidates extraction.} Following KZH13, for every anaphor %pair ($c,s$)
%()source, antecedent) pair 
we create a list of candidates by extracting all syntactic constituents from sentences which contain antecedents. Candidates that differ from antecedents in only one word, or one word and punctuation, were as well considered as antecedents. Constituents that are not antecedents are considered as negative candidates. % candidates.
%For our experiments, w

\paragraph{Data statistics.} 
	%Data statistics for both data sets is given in Table ..2.. [insert explanations from above].\\
	%\rednote{[AM: fine-tune this paragraph]}
	Table \ref{tab:data_statistics} gives statistics of the datasets: the number of anaphors (row 1), the median length (in tokens) of antecedents (row 2), the median length (in tokens) for all anaphoric sentences (row 3), the median of the number of antecedents and candidates that are not antecedents (negatives) (rows 4--5), the number of pronominal and nominal anaphors (rows 6--7). Both training sets, artificial and CSN, have only one possible antecedent for which we accept two minimal variants differing in only one word or one word and punctuation.
	%\rednote{[AM: we do not generate, if constituent parser gives us such constituents we consider them as antecedents as well, otherwise not.]} 
	%However, an anaphor in both test sets can have \bluenote{more $\rightarrow$ multiple} antecedents that differ in more than one word. 	
 On the contrary, both test sets by design allow annotation of more than one antecedent that differ in more than one word. Every anaphor in the artificial training dataset is pronominal, whereas anaphors in CSN and ASN are nominal only. ARRAU-AA has a mixture of nominal and pronominal anaphors.
\begin{table}[t]
\resizebox{0.49\textwidth}{!}{
\begin{tabular}{c|c|c|c|c|c}
\toprule
\multicolumn{2}{c}{} & \multicolumn{2}{c}{\small shell noun} &    \multicolumn{2}{c}{\small abstract anaphora} \\
\cmidrule(lr){3-4}
\cmidrule(lr){5-6}
%\multicolumn{1}{c}{} & \multicolumn{1}{c}{} &  \multicolumn{1}{c}{train} &  \multicolumn{1}{c}{test} &  \multicolumn{1}{c}{train} &  \multicolumn{1}{c}{test} \\
\multicolumn{1}{c}{} & \multicolumn{1}{c}{} &  \multicolumn{1}{c}{\makecell{\textbf{CSN} \\ {\color{darkgray} train}}} &  \multicolumn{1}{c}{\makecell{\textbf{ASN} \\ {\color{darkgray} test}}} &  \multicolumn{1}{c}{\makecell{\textbf{artifical} \\ {\color{darkgray} train}}}  &  \multicolumn{1}{c}{\makecell{\textbf{ARRAU-AA} \\ {\color{darkgray} test}}} \\
\midrule          
%\midrule                                                              
\multicolumn{2}{c|}{\# shell nouns / anaphors } &114492     & 2303  & 8527              & 600   \\
\midrule          
%\multirow{2}{*}{\parbox{0.13\linewidth}{\centering{\makecell{median \\ length}}}} & \parbox{0.22\linewidth}{\centering{antecedents}} &12.75      & 13.87 & 11                & 20.5  \\
\multirow{2}{*}{\makecell{median \\ \# of tokens}} & Antec &12.75      & 13.87 & 11                & 20.5  \\
\cmidrule{2-6}  
& AnaphS &    11.5       & 24    & 19                & 28    \\
\midrule
%\multirow{2}{*}{\parbox{0.13\linewidth}{\centering{median \#}}}  & \parbox{0.22\linewidth}{\centering{antecedents}}    & 2          & 4.5   & 2                 & 1     \\
\multirow{2}{*}{\makecell{median \\ \#}}  & Antec    & 2          & 4.5   & 2                 & 1     \\
\cmidrule{2-6}  
													& \parbox{0.22\linewidth}{\centering{negatives}}    & 44.5       & 39    & 15                & 48    \\
\midrule
\multirow{2}{*}{\#} & nominal   &114492     & 2303  & 0                 & 397   \\
\cmidrule{2-6}  
													& pronominal  & 0          & 0     & 8527              & 203   \\
													 %\cmidrule{2-6}
													%& \parbox{0.22\linewidth}{\centering{abstract}}    &-     & 485  & -      & -    \\
													%& \parbox{0.22\linewidth}{\centering{plan}}  & -     & 115  & -      & -    \\
%\midrule
%\multicolumn{2}{c|}{median distance}  & -  & 1    & -      & -   \\
\bottomrule
\end{tabular}
}
\caption{Data statistics. For the ASN and CSN we report statistics over all shell nouns, but classifiers are trained independently.}
\label{tab:data_statistics}
\vspace*{-3mm}
\end{table}

\paragraph{Data pre-processing.} Other details can be found in Supplementary Materials.

\subsection{Baselines and evaluation metrics}

%Following KZH13, predicted candidates are judged correct if they are identical to the gold annotations, admitting a single non-matching word of minor importance.\footnote{We obtained this information in personal communication with one of the authors.} 
%For instance, the following predictions are accepted as correct: this fact $\rightarrow$ \textit{\{that she is very rich}, \textit{she is very rich\}}. 
%During training all instances that differ in one word and/or punctuation from the gold antecedents are considered as positive candidates.
Following KZH13, we report {\em success@n} (\textit{s@n}), which measures whether the antecedent, or a candidate that differs in one word\footnote{We obtained this information in personal communication with one of the authors.}, is in the first $n$ ranked candidates, for $n \in \{1, 2, 3, 4\}$.\ %Henceforth 
%We will abbreviate \textit{success@n} with \textit{s@n}. 
Additionally, we report the preceding sentence baseline (PS$_{BL}$) that chooses the previous sentence for the antecedent and {\em TAGbaseline} (TAG$_{BL}$) that randomly chooses a candidate with the constituent tag label in $\{$S, VP, ROOT, SBAR$\}$.\ For TAG$_{BL}$
%baseline 
we report the average of 10 runs with 10 fixed seeds. PS$_{BL}$ always performs worse than the KZH13 model on the ASN, so we report it only for ARRAU-AA.
%
% {\em PSbaseline} (PS$_{BL}$) and {\em TAGbaseline} (TAG$_{BL}$). PS$_{BL}$ chooses the previous sentence for the antecedent,  TAG$_{BL}$
%%baseline
%randomly chooses a candidate with the constituent tag label in $\{$S, VP, ROOT, SBAR$\}$.\ For TAG$_{BL}$
%%baseline 
%we report the average of 10 runs with 10 fixed seeds. PS$_{BL}$ always performs worse than the KZH13 model on the ASN, so we report it only for ARRAU-AA.

\subsection{Training details for our models}

\textbf{Hyperparameters tuning.} We recorded performance with manually chosen HPs and then tuned HPs with Tree-structured Parzen Estimators (TPE) \cite{Bergstra2011AlgorithmsFH}\footnote{https://github.com/hyperopt/hyperopt.}. %TPE is tuned with {\em s@}1 score and $10$ trials on the devset. 
 TPE chooses HPs for the next (out of $10$) trails on the basis of the {\em s@}1 score on the devset.
As devsets we employ the ARRAU-AA corpus for shell noun resolution and the ASN corpus for unrestricted abstract anaphora resolution. For each trial we record performance on the test set. We report the best test {\em s@}1 score in $10$ trials if it is better than the scores from default HPs. %results obtained with HPs tuned with the devset. 
The default HPs and prior distributions for HPs used by TPE are given below. The (exact) HPs we used can be found in Supplementary Materials.  %\rednote{In the following we report default HPs and distributions from which TPE drew HPs.}

\textbf{Input representation.} To construct word vectors $\boldsymbol{w_i}$ as defined in Section 3, %vectors, constructed by concatenating an embedding of the word, the head of the anaphor, the context of the anaphor (average of embeddings of the anaphor, the previous and the next word), and finally, of the constituent tag of the candidate, or an embedding of the \textit{S} tag if the word is in the anaphoric sentence.
%We initialized word embeddings using $100$-dimensional GloVe vectors pre-trained on the Gigaword corpus and Wikipedia \cite{pennington2014glove}.
we used $100$-dim.\ GloVe word embeddings pre-trained on the Gigaword and Wikipedia \cite{pennington2014glove}, and did not fine-tune them. Vocabulary was built from the words in the training data with frequency in $\{3, \text{ }\mathcal{U}(1,10)\}$, and OOV words were replaced with an \textit{UNK} token. Embeddings for tags are initialized with values drawn from the uniform distribution $\mathcal{U} \big(-\frac{1}{\sqrt{d + t}}, \frac{1}{\sqrt{d+t}}\big)$, where $t$ is the number of tags\footnote{We used a list of tags obtained from the Stanford Parser.} and $d \in \{50, \textit{qlog-}\mathcal{U}(30,100)\}$ the size of the tag embeddings.\footnote{\textit{qlog-}$\mathcal{U}$ is the so-called qlog-uniform distribution.} %Cf.\ 
%See	https://github.com/hyperopt/hyperopt/wiki/FMin.}
%for more details.} 
We experimented with removing embeddings for tag, anaphor and context. 

\textbf{Weights initialization.} The size of the LSTMs hidden states was set to $\{100,$ $\textit{qlog-}\mathcal{U}(30,150) \}$. We initialized the weight matrices of the LSTMs with random orthogonal matrices \cite{Henaff2016RecurrentON}, all other weight matrices with the initialization proposed in \citet{He2015DelvingDI}. The first feed-forward layer size is set to a value in $\{400, \textit{qlog-}\mathcal{U}(200,800)\}$, the second to a value in $\{1024, \textit{qlog-}\mathcal{U}(400,2000)\}$. Forget biases in the LSTM were initialized with 1s \cite{Jzefowicz2015AnEE}, all other biases with 0s.

\textbf{Optimization.} We trained our model in mini-batches using Adam
%optimizer 
\cite{kingma2014adam} with 
%the 
the learning rate of $10^{-4}$
%. %The
%M
and maximal batch size
%\rednote{was}
% set to 
$64$.
% MOVED TO SUPPLEMENT
% \bluenote{PRE-PROCESSING?: All sentences in the batch are padded with a \textit{PAD} token up to the maximal sentence length in the batch and corresponding hidden states in the LSTM are masked with zeros. To implement the model efficiently in TensorFlow\footnote{The code and the data will be available.} \cite{tensorflow2015-whitepaper}, batches are constructed such that every sentence instance in the batch has the same number of positive candidates and the same number of negative candidates.\footnote{Note that by this we do \textbf{not} mean that the ratio of positive and negative examples is 1:1.}} 
We clip gradients by global norm \cite{Pascanu2013OnTD}, with 
a 
clipping value in $\{1.0, \text{ }\mathcal{U}(1,100)\}$. We train 
%the model 
for $10$ epochs and choose the model 
%which
that performs best on
the devset. %If the training {\em s@}1 converges to $99.99$, we stop the training. Depending on the training data size one epoch takes around 2-8 minutes using one Nvidia GeForce GTX1080 gpu.

\textbf{Regularization.} We used the 
$l_2$-regularization with $\lambda \in \{10^{-5}, \textit{ log-}\mathcal{U}(10^{-7},10^{-2})\}$. Drop\-out \cite{Srivastava2014DropoutAS} with a 
keep pro\-ba\-bi\-li\-ty  $k_p\in\{0.8, \text{ } \mathcal{U}(0.5,1.0)\}$ was applied to the outputs of the LSTMs, both feed-forward layers
and optionally to the input with $k_p \in \mathcal{U}(0.8,1.0)$.

	\section{Results and analysis}

\begin{table}[t]
\center
\resizebox{0.49\textwidth}{!}{
\begin{tabular}{c|c|cccc}
\toprule
\multicolumn{1}{r}{} & \multicolumn{1}{r}{}& \multicolumn{1}{r}{s @ 1}& \multicolumn{1}{r}{s @ 2} & \multicolumn{1}{r}{s @ 3} & \multicolumn{1}{r}{s @ 4} \\
\midrule
\multirow{3}{*}{\parbox{0.28\linewidth}{\centering \textbf{fact}\\ (train: 43809,\\ test: 472)}}        & MR-LSTM & \textbf{83.47} & \textbf{85.38} & \textbf{86.44} & \textbf{87.08} \\
                             														 & KZH13        & 70.00 & 86.00 & 92.00 & 95.00 \\
                             														 & TAG$_{BL}$ & 46.99 & -     & -     & -     \\
\midrule														 
\multirow{4}{*}{\parbox{0.28\linewidth}{\centering \textbf{reason}\\ (train: 4529,\\ test: 442)}}      & MR-LSTM     & 71.27 & 77.38 & 80.09 & 80.54 \\
																	   & + tuning      & \textbf{87.78} & \textbf{91.63}          & \textbf{93.44} & \textbf{93.89} \\
																	   & KZH13        & 72.00 & 86.90 & 90.00 & 94.00 \\
																	   & TAG$_{BL}$ & 42.40 & -     & -     & -     \\
\midrule																	   
 \multirow{3}{*}{\parbox{0.28\linewidth}{\centering \textbf{issue}\\ (train: 2664,\\ test: 303)}}       & MR-LSTM     & \textbf{88.12} & \textbf{91.09} & \textbf{93.07} & \textbf{93.40} \\
																	  & KZH13        & 47.00 & 61.00 & 72.00 & 81.00 \\
																	  & TAG$_{BL}$ & 44.92 & -     & -     & -     \\
\midrule																	  
 \multirow{3}{*}{\parbox{0.28\linewidth}{\centering \textbf{decision} \\(train: 42289,\\ test: 389)}}    & MR-LSTM     & \textbf{76.09} & \textbf{85.86} & \textbf{91.00} & \textbf{93.06} \\
																	  & KZH13        & 35.00 & 53.00 & 67.00 & 76.00 \\
																	  & TAG$_{BL}$ & 45.55 & -     & -     & -     \\
\midrule																	  
 \multirow{3}{*}{\parbox{0.28\linewidth}{\centering \textbf{question}\\ (train: 9327,\\ test: 440)}}      & MR-LSTM     & \textbf{89.77} & \textbf{94.09} & \textbf{95.00} & \textbf{95.68} \\
																	  & KZH13        & 70.00 & 83.00 & 88.00 & 91.00 \\
																	  & TAG$_{BL}$ & 42.02 & -     & -     & -     \\
\midrule																	  
 \multirow{3}{*}{\parbox{0.28\linewidth}{\centering \textbf{possibility} (train: 11874, test: 277)}} & MR-LSTM     & \textbf{93.14} & \textbf{94.58} & \textbf{95.31} & \textbf{95.67} \\
																	  & KZH13        & 56.00 & 76.00 & 87.00 & 92.00 \\
																	  & TAG$_{BL}$ & 48.66 & -     & -     & -\\                              
\bottomrule
\end{tabular}
}
\caption{Shell noun resolution results. }\label{tab:results_asn}
\end{table}

\subsection{Results on shell noun resolution dataset}
	%Comparison with published results on shell noun resolution}

%Table \ref{tab:results_asn} shows model performance on the ASN corpus without tuning of HPs, unless stated otherwise.
Table \ref{tab:results_asn} provides the results of the mention-ranking model (MR-LSTM) on the ASN corpus using default HPs. Column 2 states which model produced the results: KZH13 refers to the best reported results in \citet{kolhatkar-zinsmeister-hirst:2013:EMNLP} and TAG$_{BL}$ is the baseline described in Section 5.2.   
%\bluenote{AF: you say above that KZH13 is always better than PSBL, which is therefore not given. But you don't say how much better. This might be informative; add the PSBL for ASN, too?} \rednote{AM: don't have the info about the distance in jsons. Could ask Juri to calculate it.} \bluenote{AF: no high priority}
%Table 3 shows that the MR-LSTM without any tuning outperforms KZH13's results by large margins for 5/6 shell nouns in terms of  {\em s@1} score.%, for all ranks $n$ of {\em s@n}. 

In terms of \textit{s@}1 score, MR-LSTM outperforms both KZH13's results and TAG$_{BL}$ without even necessitating HP tuning. For the outlier \textit{reason} %To investigate whether the choice of HPs or the architecture are causes of the weaker results for \textit{reason}, 
we tuned HPs (on ARRAU-AA) for different variants of the architecture: the full architecture, without embedding of the context of the anaphor (ctx), of the anaphor (aa), of both constituent tag embedding and shortcut (tag,cut), dropping only the shortcut (cut), using only word embeddings as input 
%to the bi-LSTM
(ctx,aa,tag,cut),  without the first (ffl1) and second (ffl2) layer. From Table \ref{tab:architecture_asn} we observe:
%shows the performances.
% of these variants. 
%We observe: 
(1) %the choice of default HPs was not optimal for \textit{reason},
with HPs tuned on ARRAU-AA, we obtain results well beyond KZH13, (2) all ablated model variants perform worse than the full model, (3) a large performance drop when omitting syntactic information (tag,cut) suggests that the model makes good use of it. However, this could also be due to a
%mean that the model exploits a 
bias in the tag distribution, given that all candidates %constituents 
stem from the single sentence that contains antecedents. The median occurrence of the S tag among both antecedents and negative candidates is $1$, thus the model could achieve $50.00$ \textit{s@}1 by picking S-type constituents, just as TAG$_{BL}$ achieves $42.02$ for \textit{reason} and 
%even 
$48.66$ for \textit{possibility}.
%We also notice that it was 
%We also tuned HPs to give 
%meaningful 

Tuning of HPs gives us insight into how different model variants cope with the task. For example, without tuning the model with and without syntactic information achieves $71.27$ and $19.68$ (not shown in table) \textit{s@}1 score, respectively, and with tuning: $87.78$ and $68.10$. Performance of $68.10$ \textit{s@}1 score indicates that the model is able to learn 
%to some degree 
without syntactic guidance, contrary to the $19.68$ \textit{s@}1 score before tuning. 
\begin{table}[t]
\center
\resizebox{0.49\textwidth}{!}{
\begin{tabular}{c|c|c|c|c|c|cccc}
\toprule 
\multicolumn{10}{c}{\textbf{reason}} \\ 
ctx & aa & tag & cut &  ffl1 & ffl2 & s@1 & s@2 & s@ 3 & s@ 4  \\
\midrule    
{\color{darkgreen} \cmark} & {\color{darkgreen} \cmark} & {\color{darkgreen} \cmark} & {\color{darkgreen} \cmark} & {\color{darkgreen} \cmark} & {\color{darkgreen} \cmark} &  \textbf{87.78} & \textbf{91.63} & \textbf{93.44} & \textbf{93.89} \\
{\color{darkred} \xmark} & {\color{darkgreen} \cmark} & {\color{darkgreen} \cmark} & {\color{darkgreen} \cmark} & {\color{darkgreen} \cmark} & {\color{darkgreen} \cmark} & 85.97 & 87.56 & 89.14 & 89.82 \\
{\color{darkgreen} \cmark} & {\color{darkred} \xmark} & {\color{darkgreen} \cmark} & {\color{darkgreen} \cmark} & {\color{darkgreen} \cmark} & {\color{darkgreen} \cmark} & 86.65 & 88.91 & 91.18 & 91.40 \\
{\color{darkgreen} \cmark} & {\color{darkgreen} \cmark} & {\color{darkred} \xmark} & {\color{darkred} \xmark} & {\color{darkgreen} \cmark} & {\color{darkgreen} \cmark} &  68.10 & 80.32 & 85.29 & 89.37 \\
{\color{darkgreen} \cmark} & {\color{darkgreen} \cmark} & {\color{darkgreen} \cmark} & {\color{darkred} \xmark} & {\color{darkgreen} \cmark} & {\color{darkgreen} \cmark} &  85.52 & 88.24 & 89.59 & 90.05 \\
{\color{darkred} \xmark} & {\color{darkred} \xmark} & {\color{darkred} \xmark} & {\color{darkred} \xmark} & {\color{darkgreen} \cmark} & {\color{darkgreen} \cmark} & 66.97 & 80.54 & 85.75 & 88.24 \\
{\color{darkgreen} \cmark} & {\color{darkgreen} \cmark} & {\color{darkgreen} \cmark} & {\color{darkgreen} \cmark} & {\color{darkred} \xmark} & {\color{darkgreen} \cmark} &  87.56 & 91.63 & 92.76 & 94.12\\
{\color{darkgreen} \cmark} & {\color{darkgreen} \cmark} & {\color{darkgreen} \cmark} & {\color{darkgreen} \cmark} & {\color{darkgreen} \cmark} & {\color{darkred} \xmark} &  85.97 & 88.69 & 89.14 & 90.05 \\
\bottomrule
\end{tabular}
}
\caption{Architecture ablation for \textit{reason}.}\label{tab:architecture_asn}
\vspace*{-3mm}
\end{table}

\begin{table*}[t]
\center
\resizebox{0.98\textwidth}{!}{
\begin{tabular}{c|c|c|c|c|c|cccc|cccc|cccc}
\toprule 
\multicolumn{6}{c}{} & \multicolumn{4}{c}{\textbf{all (600)}} & \multicolumn{4}{c}{\textbf{nominal (397)}} & \multicolumn{4}{c}{\textbf{pronominal (203)}} \\ 
ctx & aa & tag & cut &  ffl1 & ffl2 & s@1 & s@2 & s@ 3 & s@ 4 & s@1 & s@2 & s@ 3 & s@ 4 & s@1 & s@2 & s@ 3 & s@ 4 \\
\midrule    
{\color{darkgreen} \cmark} & {\color{darkgreen} \cmark} & {\color{darkgreen} \cmark} & {\color{darkgreen} \cmark} & {\color{darkgreen} \cmark} & {\color{darkgreen} \cmark} &  24.17 & 43.67 & 54.50 & 63.00 & 29.47 & 50.63 & 62.47 & 72.04 & 13.79 & 30.05 & 38.92 & 45.32 \\
{\color{darkred} \xmark} & {\color{darkgreen} \cmark} & {\color{darkgreen} \cmark} & {\color{darkgreen} \cmark} & {\color{darkgreen} \cmark} & {\color{darkgreen} \cmark} & 29.67 & 52.50 & 66.00 & \textbf{75.00} & 33.50 & 58.19 & 72.04 & \textbf{80.86} & 22.17 & 41.38 & \textbf{54.19} & \textbf{63.55} \\
{\color{darkgreen} \cmark} & {\color{darkred} \xmark} & {\color{darkgreen} \cmark} & {\color{darkgreen} \cmark} & {\color{darkgreen} \cmark} & {\color{darkgreen} \cmark} &22.83 & 39.00 & 52.00 & 61.33 & 22.42 & 41.31 & 54.66 & 64.48 & 23.65 & 34.48 & 46.80 & 55.17 \\
{\color{darkgreen} \cmark} & {\color{darkgreen} \cmark} & {\color{darkred} \xmark} & {\color{darkred} \xmark} & {\color{darkgreen} \cmark} & {\color{darkgreen} \cmark} &  38.33 & 54.83 & 63.17 & 69.33 & 46.60 & \textbf{64.48} & 72.54 & 79.09 & 22.17 & 35.96 & 44.83 & 50.25 \\
{\color{darkgreen} \cmark} & {\color{darkgreen} \cmark} & {\color{darkred} \xmark} & {\color{darkred} \xmark} & {\color{darkgreen} \cmark} & {\color{darkgreen} \cmark} & \textbf{43.83} & \textbf{56.33} & \textbf{66.33} & 73.00 & \textbf{51.89} & \textbf{64.48} & \textbf{73.55} & 79.85 & 28.08 & 40.39 & 52.22 & 59.61\\
{\color{darkgreen} \cmark} & {\color{darkgreen} \cmark} & {\color{darkred} \xmark} & {\color{darkred} \xmark} & {\color{darkgreen} \cmark} & {\color{darkgreen} \cmark} & 38.17 & 52.50 & 61.33 & 68.67 & 43.07 & 57.43 & 65.49 & 72.04 & 28.57 & \textbf{42.86} & 53.20 & 62.07\\
{\color{darkgreen} \cmark} & {\color{darkgreen} \cmark} & {\color{darkgreen} \cmark} & {\color{darkred} \xmark} & {\color{darkgreen} \cmark} & {\color{darkgreen} \cmark} &30.17 & 48.00 & 57.83 & 67.33 & 30.73 & 50.88 & 61.21 & 71.54 & \textbf{29.06} & 42.36 & 51.23 & 59.11 \\
{\color{darkred} \xmark} & {\color{darkred} \xmark} & {\color{darkred} \xmark} & {\color{darkred} \xmark} & {\color{darkgreen} \cmark} & {\color{darkgreen} \cmark} & 26.33 & 40.50 & 50.67 & 58.67 & 28.46 & 41.81 & 52.14 & 59.70 & 22.17 & 37.93 & 47.78 & 56.65 \\
{\color{darkgreen} \cmark} & {\color{darkgreen} \cmark} & {\color{darkgreen} \cmark} & {\color{darkgreen} \cmark} & {\color{darkred} \xmark} & {\color{darkgreen} \cmark} &  21.33 & 41.17 & 53.17 & 60.33 & 23.43 & 47.36 & 60.45 & 69.52 & 17.24 & 29.06 & 38.92 & 42.36 \\
{\color{darkgreen} \cmark} & {\color{darkgreen} \cmark} & {\color{darkgreen} \cmark} & {\color{darkgreen} \cmark} & {\color{darkgreen} \cmark} & {\color{darkred} \xmark} &  12.00 & 24.67 & 33.50 & 41.50 & 13.35 & 27.20 & 37.28 & 45.84 & 9.36  & 19.70 & 26.11 & 33.00 \\
\midrule  
\multicolumn{6}{c|}{PS$_{BL}$} &  27.67   &  -     &    -   &     -  &30.48   &   -    &    -   &    -   & 22.17   &    -   &   -    &     -  \\
\multicolumn{6}{c|}{TAG$_{BL}$} &  38.43& -     & -     & -     & 40.10 & -     & -     & -     & \textbf{35.17} & -     & -     & - \\
\bottomrule
\end{tabular}
}
\caption{Results table for the ARRAU-AA test set. Refer to text for explanation of duplicated rows.%Results in the row 4 are obtained with HPs which yielded the best s@1 score for all anaphors and in the row 6 for pronominal anaphors. Results in rows 4--5 are obtained with the same HPs, when training data is shuffled (row 5) and not (row 4).
}\label{tab:results_arrau}
\end{table*}

\subsection{Results on the ARRAU corpus}

Table \ref{tab:results_arrau} shows the performance of different variants of the MR-LSTM with HPs tuned on the ASN corpus (always better than the default HPs), when evaluated on $3$ different subparts of the ARRAU-AA: all $600$ abstract anaphors, $397$ nominal and $203$ pronominal ones. %For simplicity, we will refer to abstract (pro)nominal anaphora as (pro)nominal anaphora.
HPs were tuned on the ASN corpus for every variant separately, without shuffling of the training data. For the best performing variant, without syntactic information (tag,cut), we report the results with HPs that yielded the best \textit{s@}1 test score for all anaphors (row 4), when training with those HPs on shuffled training data (row 5), and with HPs that yielded the best \textit{s@}1 score for pronominal anaphors (row 6).

%The MR-LSTM is more successful in resolving nominal anaphors than pronominal, although the training data provides only pronominal anaphors. This indicates that resolving pronominal
%%(any type of)
%abstract anaphora is harder compared to nominal abstract anaphora, such as shell nouns. Moreover, for shell noun resolution in KZH13's dataset, the MR-LSTM achieved \textit{s@}1 scores in the range of $76.09$ -- $93.14$, while the best variant of the model achieves $51.89$ \textit{s@}1 score for resolving nominal anaphors in ARRAU-AA. Although lower performance is expected, since we do not have specific
%%separate 
%training data for individual nominal anaphors in ARRAU-AA, we suspect that the reason for better performance for shell noun resolution in KZH13 is due to a larger number of positive candidates in ASN (cf.\ Table \ref{tab:data_statistics}, rows: pos/neg). 
%

The MR-LSTM is more successful in resolving nominal than pronominal anaphors, although the training data provides only pronominal ones. This indicates that resolving pronominal
%(any type of)
abstract anaphora is harder compared to nominal abstract anaphora, such as shell nouns. Moreover, for shell noun resolution in KZH13's dataset, the MR-LSTM achieved \textit{s@}1 scores in the range $76.09$--$93.14$, while the best variant of the model achieves $51.89$ \textit{s@}1 score for nominal anaphors in ARRAU-AA. Although lower performance is expected, since we do not have specific
%separate
training data for individual nominals %\rednote{(but only pronominals)}
%anaphors
in ARRAU-AA, we suspect that the reason for better performance for shell noun resolution in KZH13 is due to a larger number of positive candidates in ASN (cf.\ Table \ref{tab:data_statistics}, rows: antecedents/negatives). 

We also note that HPs that yield good performance for resolving nominal anaphors are not necessarily good for pronominal ones (cf.\ rows 4--6 in Table \ref{tab:results_arrau}). Since the TPE tuner was tuned on the nominal-only ASN data, this suggest that it would be better to tune HPs for pronominal anaphors on a different dataset or stripping the nouns in ASN. 

Contrary to %what we observed for shell noun evaluation on ASN
shell noun resolution, omitting syntactic information boosts performance in ARRAU-AA. We conclude that when the model is provided with syntactic information, it learns to pick S-type candidates, but does not continue to learn deeper features to further distinguish them or needs more data to do so. Thus,
%For this reason, 
the model is not able to point to exactly one antecedent, resulting  in a lower \textit{s@}1 score, but does well in picking a few good candidates, which yields good \textit{s@}2-4 scores. This is what we can observe
%We observe exactly that 
from row 2 vs.\ row 6 in Table \ref{tab:results_arrau}: 
%The variant of the
the MR-LSTM without context embedding (ctx) achieves a comparable \textit{s@}2 score with the variant that omits syntactic information, but better \textit{s@}3-4 scores. Further, median occurrence of tags not in \{S, VP, ROOT, SBAR\} among top-4 ranked candidates is $0$ for the full architecture, and $1$ when syntactic information is omitted. The need for discriminating capacity
%property 
of the model is more emphasized in ARRAU-AA, given that the median occurrence of S-type candidates among negatives is $2$ for nominal and even $3$ for pronominal anaphors, whereas it is $1$ for ASN.
%is $2$ and of pronominal even $3$. 
This is in line with the lower TAG$_{BL}$ in ARRAU-AA. 

Finally, not all parts of the architecture contribute to system performance,
%to the overall architecture
contrary to what is observed for
%it was the case 
\textit{reason}. For nominal anaphors, 
%embedding of 
the anaphor (aa) and feed-forward layers (ffl1, ffl2) are beneficial, for pronominals only the second
%it is only the second feed-forward layer. % 
ffl.

\subsection{Exploring the model}

We finally analyze deeper aspects of the model: (1) whether a learned representation between the anaphoric {\em sentence} and an antecedent establishes a relation between {\em a specific anaphor we want to resolve} and the antecedent and (2) whether the max-margin objective enforces a separation of the joint representations in the shared space. 

\begin{figure}[t]
\centering
\includegraphics[width=0.85\linewidth]{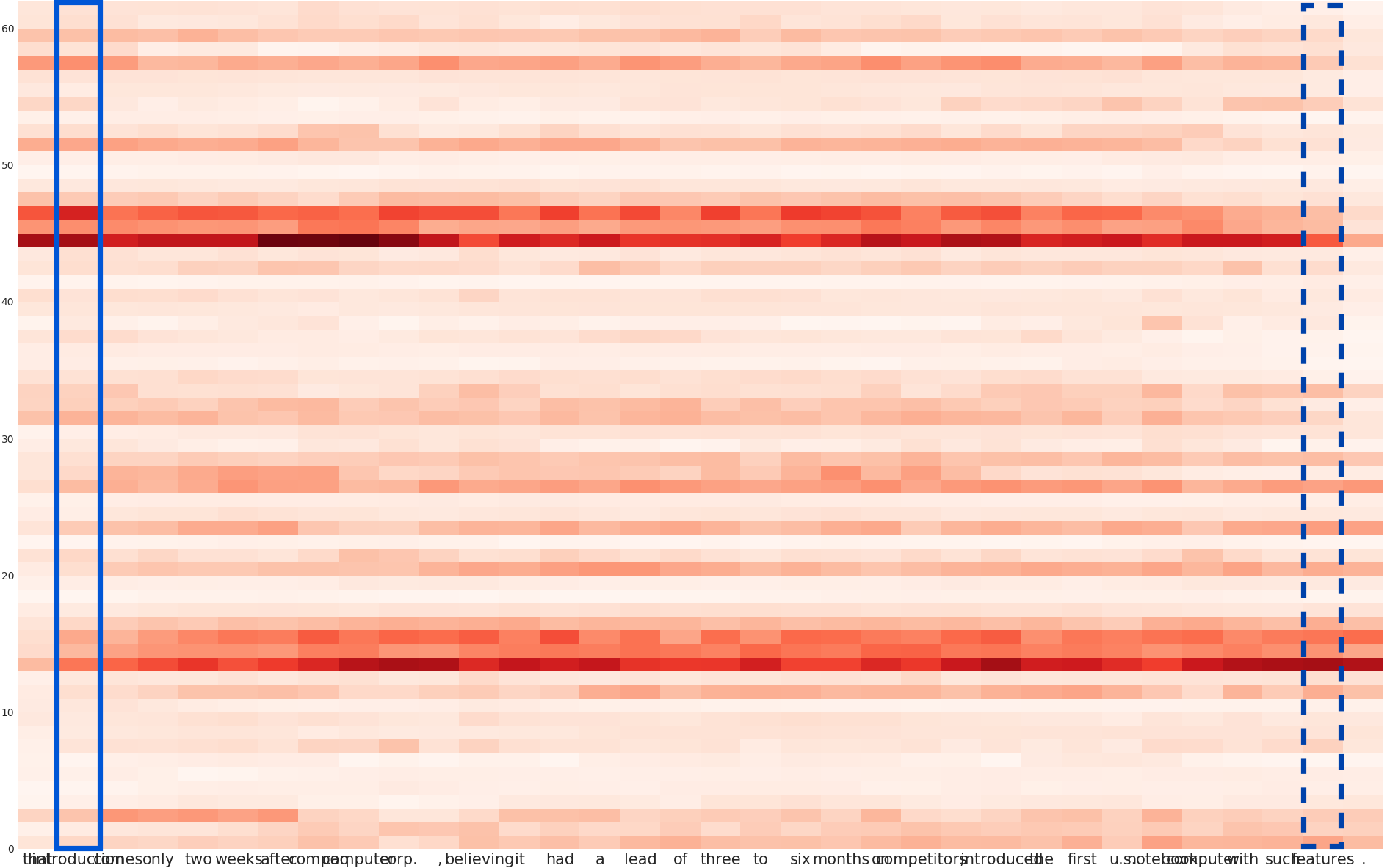}
\caption{Visualizing the differences between outputs of the bi-LSTM over time for an anaphoric sentence containing two anaphors.}
\label{fig:heatmap}
\vspace*{-4mm}
\end{figure}

(1) 
We claim that 
by providing embeddings of both the anaphor and the sentence containing the anaphor we ensure that the learned relation between antecedent and anaphoric sentence
%(possibly containing several anaphors), 
is dependent on the anaphor under consideration. %Consider following sentence with two anaphors in bold: \textit{\textbf{That introduction} comes only two weeks after Compaq Computer Corp.\, believing it had a lead of three to six months on competitors, introduced the first U.S.\ notebook computer with \textbf{such features}}.
Fig.\ \ref{fig:heatmap} illustrates the heatmap for an anaphoric sentence with two anaphors. The i-th column of the heatmap corresponds to absolute differences between the output of the bi-LSTM for the i-th word in the anaphoric sentence when the first vs.\ second anaphor is resolved.
%, and the output of the bi-LSTM for the i-th word in the anaphoric sentence when the second anaphor is under consideration. 
Stronger color indicates larger difference, the blue rectangle represents the column for the head of the first anaphor, the dashed blue rectangle the column for the head of the second anaphor. Clearly, the representations differ when the first vs.\  second anaphor is being resolved and 
consequently, joint representations with an antecedent will differ too. 

\begin{figure}[t]
	\centering
	\includegraphics[width=0.85\linewidth]{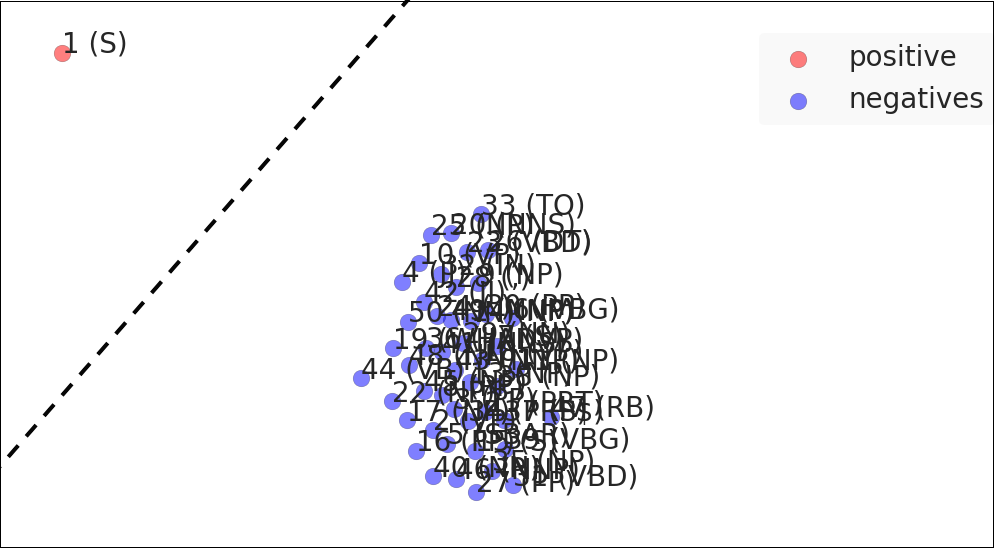}
	\caption{tSNE projection of outputs of ffl2. Labels are the predicted ranks and the constituent tag.}
	\label{fig:tsne}
	\vspace*{-3mm}
\end{figure}

(2) It is known that the max-margin objective separates the best-scoring positive candidate from the best-scoring negative candidate. To investigate what the objective accomplishes in the MR-LSTM model, we analyze
%investigated
%what happens with 
the joint representations of candidates and the anaphoric sentence (i.e., outputs of ffl2)
%the second feed-forward layer)
 after training. For a randomly chosen instance from ARRAU-AA, we plotted outputs of 
 %the second feed-forward layer 
 ffl2 %projected in a 2-D space
 with the tSNE algorithm \cite{van2008visualizing}. %From Figure \ref{fig:tsne} we can see 
 Fig.\ \ref{fig:tsne} illustrates that the joint representation of %the antecedent (the first ranked candidate) 
 the first ranked candidate and the anaphoric sentence is clearly separated from %all 
 other joint representations. 
%Meaning 
This shows that the max-margin objective separates the best scoring positive candidate from the best scoring negative candidate by separating their respective joint representations with the anaphoric sentence.

%\begin{figure}[t]
%	\centering
%	\includegraphics[width=0.85\linewidth]{figs/joint.png}
%	\caption{tSNE projection of outputs of ffl2. Labels are the predicted ranks and the constituent tag.}
%	\label{fig:tsne}
%	\vspace*{-3mm}
%\end{figure}
%\begin{figure}
%\centering
%    \begin{subfigure}{0.48\textwidth}
%        \centering
%        \includegraphics[width=\linewidth]{figs/heatmap_0.png}
%        %\caption{}\label{fig:hm0}
%    \end{subfigure} %
%    \begin{subfigure}{0.48\textwidth}
%        \centering
%        \vspace{0.2cm}
%        \includegraphics[width=\linewidth]{figs/heatmap_1.png}
%        %\caption{}\label{fig:hm1}
%    \end{subfigure} 

%when the constituent tag embedding and a shortcut for the TAG information are omitted from the input (-TAG \& shortcut), when the anaphora embedding is omitted (-PA), when only the word embedding serves as the input (-TAG \& shortcut \&PA), when we leave the constituent embeddings in the input, but drop the shortcut (-shortcut), and removing feedforward networks from the architecture (-FFNNs).
	\section{Conclusions}

We presented a neural mention-ranking model for the resolution of unconstrained abstract anaphora, and applied it to two datasets with different types of abstract anaphora: the shell noun dataset and a subpart of ARRAU with (pro)nominal abstract anaphora of any type.
To our knowledge this work is the first to address the unrestricted abstract anaphora resolution task with a neural network. Our model also outperforms state-of-the-art results %of prior work 
on the shell noun dataset. 

In this work we explored the use of purely artificially created training data and how far it can bring us. 
%, and in
In future work, we plan to
% we may also 
investigate mixtures of (more) artificial and natural data from different sources (e.g.\ ASN, CSN). 

On the more challenging ARRAU-AA, %our model
we found model variants that surpass the baselines for %resolving nominal anaphors and across types,
the entire and the nominal part of ARRAU-AA, although we do not train models on individual (nominal) anaphor training data like the related work for shell noun resolution. %, while pronominal ones lag behind. 
However, our model still lags behind for pronominal anaphors. Our results suggest that models for nominal and pronominal anaphors should be learned independently, starting with tuning of HPs on a more suitable devset for pronominal anaphors. %Refining models for the types separately is left for future work.

We show that the model can exploit syntactic information %about the constituent tag of a candidate 
to select plausible candidates, but that when it does so, it does not learn how to distinguish candidates of equal syntactic type.
%, \rednote{or needs more data to do so}. 
By contrast, if the model is not provided with syntactic information, 
%about the constituent tag 
it learns deeper features that enable it to pick the correct antecedent without narrowing down the choice of candidates. Thus, in order to improve performance, the model should be enforced to first select reasonable candidates and then continue to learn features to distinguish them, using a larger training set that is easy to provide.

%Our next step for future work will be to design such a model, and to offer it antecedent candidates chosen not only from the sentence that contains the antecedent, but from a larger context. 
In future work we will design such a model, and offer it candidates chosen not only from sentences containing the antecedent, but the larger context.

	\section*{Acknowledgments}

This work has been supported by the German Research Foundation as part of the Research Training Group Adaptive Preparation of Information from Heterogeneous Sources (AIPHES) under grant No.\ GRK 1994/1.
We would like to thank anonymous reviewers for useful comments and especially thank Todor Mihaylov for the model implementations advices and everyone in the Computational Linguistics Group for helpful discussion.
	\bibliography{acl2017}
	\bibliographystyle{emnlp_natbib}
	
	\appendix
	\section{Pre-processing details}
The CSN corpus we obtained from the authors contained tokenized sentences for antecedents and anaphoric sentences. The number of instances differed from the reported numbers in KZH13 in $9$ to $809$ instances for training, and $1$ for testing. %(cf. Table \ref{tab:results_asn}). %\citet{kolhatkar-zinsmeister-hirst:2013:EMNLP}.
%Additionally
The given 
%anaphoric 
sentences still contained the antecedent, so we removed it from the sentence and transformed the corresponding shell %noun was transformed 
into "\textit{this} $\langle \textit{shell noun} \rangle$". 
%For example, following transformation was done: 
An example of this process is:
\textit{The decision \textbf{to disconnect the ventilator} came after doctors found no brain activity.}\ $\rightarrow$ \textit{This decision came after doctors found no brain activity.}

To use pre-trained word embeddings we had to lowercase all the data. As we use an automatic parse to extract all syntactic constituents, due to parser errors, candidates with the same string appeared with different tags. We eliminated duplicates by checking which tag is more frequent for candidates which have the same POS tag of the first word as the duplicated candidate, in the whole dataset. In case duplicated candidates were still occurring, we chose any of them. If such duplicates occur in antecedents, we don't take such instances in the training data to eliminate noise, or choose any of them for the test data. For the training data we choose instances with an anaphoric sentence length of at least $10$ tokens. 

All sentences in the batch are padded with a \textit{PAD} token up to the maximal sentence length in the batch and corresponding hidden states in the LSTM are masked with zeros. To implement the model efficiently in TensorFlow, batches are constructed in such a way that every sentence instance in the batch has the same number of positive candidates and the same number of negative candidates. Note that by this we do \textbf{not} mean that the ratio of positive and negative examples is 1:1.

\begin{table*}[t]
\center
\resizebox{\textwidth}{!}{
\begin{tabular}{c|c|c|c|c|c|ccccccccccccc}
\toprule 
ctx & aa & tag & cut &  ffl1 & ffl2 & $h_{LSTM}$ & $h_{ffl1}$ & $h_{ffl2}$ & $d_{TAG}$ & $g$ & $f_w$ & $r$ & $k_{LSTM}$ & $k_{ffl1}$ & $k_{ffl2}$ & \# param. & $t_e$ & $e$ \\
\midrule    
{\color{darkgreen} \cmark} & {\color{darkgreen} \cmark} & {\color{darkgreen} \cmark} & {\color{darkgreen} \cmark} & {\color{darkgreen} \cmark} & {\color{darkgreen} \cmark} & 95          & 283              & 1115             & 49            & 2.13      & 9.40       & $6.61^{-5}$       & 0.60                  & 0.99           & 0.71           & 1928557    & 3.86               & 9     \\
{\color{darkred} \xmark} & {\color{darkgreen} \cmark} & {\color{darkgreen} \cmark} & {\color{darkgreen} \cmark} & {\color{darkgreen} \cmark} & {\color{darkgreen} \cmark} & 140         & 375              & 1193             & 83            & 7.44      & 4.41       &  $2.87^{-6}$       & 0.62                  & 0.80           & 0.82           & 2842489    & 3.83               & 5     \\
{\color{darkgreen} \cmark} & {\color{darkred} \xmark} & {\color{darkgreen} \cmark} & {\color{darkgreen} \cmark} & {\color{darkgreen} \cmark} & {\color{darkgreen} \cmark} & 61          & 621              & 1485             & 81            & 8.27      & 3.27       & $3.44^{-3}$       & 0.56                  & 0.94           & 0.99           & 3502713    & 3.71               & 6     \\
{\color{darkgreen} \cmark} & {\color{darkgreen} \cmark} & {\color{darkred} \xmark} & {\color{darkred} \xmark} & {\color{darkgreen} \cmark} & {\color{darkgreen} \cmark} & 39          & 722              & 1655             & -             & 43.00     & 7.11       & $2.91^{-6}$       & 0.89                  & 0.99           & 0.88           & 3624949    & 3.73               & 1     \\
{\color{darkgreen} \cmark} & {\color{darkgreen} \cmark} & {\color{darkgreen} \cmark} & {\color{darkred} \xmark} & {\color{darkgreen} \cmark} & {\color{darkgreen} \cmark} &  79          & 359              & 1454             & 65            & 3.22      & 7.74       & $9.66^{-6}$       & 0.70                  & 0.76           & 0.94           & 2459362    & 4.61               & 2     \\
{\color{darkred} \xmark} & {\color{darkred} \xmark} & {\color{darkred} \xmark} & {\color{darkred} \xmark} & {\color{darkgreen} \cmark} & {\color{darkgreen} \cmark} &38          & 548              & 1997             & -             & 82.00     & 7.14       &  $6.07^{-3}$ & 0.52                  & 0.98           & 0.80           & 3345859    & 4.41               & 4     \\
{\color{darkgreen} \cmark} & {\color{darkgreen} \cmark} & {\color{darkgreen} \cmark} & {\color{darkgreen} \cmark} & {\color{darkred} \xmark} & {\color{darkgreen} \cmark} &  39          & -                & 956              & 96            & 7.82      & 8.68       & $1.64^{-7}$       & 0.78                  & -              & 0.59           & 1567647    & 4.41               & 3     \\
{\color{darkgreen} \cmark} & {\color{darkgreen} \cmark} & {\color{darkgreen} \cmark} & {\color{darkgreen} \cmark} & {\color{darkgreen} \cmark} & {\color{darkred} \xmark} & 71          & 305              & -                & 94            & 9.40      & 5.42       & $8.3^{-3}$       & 0.52                  & 0.83           & -              & 1593880    & 4.62               & 8    \\
\bottomrule
\end{tabular}
}
\caption{HPs used for the different architecture variants for the shell noun \textit{reason}.}
\label{tab:hp1}

\center
\resizebox{\textwidth}{!}{
\begin{tabular}{c|c|c|c|c|c|cccccccccccccc}
\toprule 
ctx & aa & tag & cut &  ffl1 & ffl1 &  $h_{LSTM}$ & $h_{ffl1}$ & $h_{ffl2}$ & $d_{TAG}$ & $g$ & $f_w$ & $r$  & $k_{LSTM}$ & $k_{input}$ & $k_{ffl1}$ & $k_{ffl2}$ & \# param. & $t_e$ & $e$ \\
\midrule    
{\color{darkgreen} \cmark} & {\color{darkgreen} \cmark} & {\color{darkgreen} \cmark} & {\color{darkgreen} \cmark} & {\color{darkgreen} \cmark} & {\color{darkgreen} \cmark} &  37  & 684 & 1081 & 99  & 7.40  & 2.41 & $2.38^{-4}$        & 0.58 & 0.86 & 0.70 & 0.96 & 3716655 & 2.69 & 2   \\

{\color{darkred} \xmark} & {\color{darkgreen} \cmark} & {\color{darkgreen} \cmark} & {\color{darkgreen} \cmark} & {\color{darkgreen} \cmark} & {\color{darkgreen} \cmark} &59  & 520 & 520  & 71  & 3.06  & 3.59 & $2.54^{-4}$       & 0.70 & 0.83 & 0.56 & 0.89 & 2592937 & 2.62 & 1   \\

{\color{darkgreen} \cmark} & {\color{darkred} \xmark} & {\color{darkgreen} \cmark} & {\color{darkgreen} \cmark} & {\color{darkgreen} \cmark} & {\color{darkgreen} \cmark} &45  & 782 & 447  & 31  & 1.50  & 6.20 & $1.22^{-4}$ & 0.85 & 0.90 & 0.52 & 0.87 & 2300531 & 2.62 & 1   \\

{\color{darkgreen} \cmark} & {\color{darkgreen} \cmark} & {\color{darkred} \xmark} & {\color{darkred} \xmark} & {\color{darkgreen} \cmark} & {\color{darkgreen} \cmark} &  36  & 423 & 417  & -   & 46.00 & 3.64 & $1.75^{-5}$        & 0.57 & 0.86 & 0.65 & 0.63 & 2271652 & 8.24 & 2   \\
{\color{darkgreen} \cmark} & {\color{darkgreen} \cmark} & {\color{darkred} \xmark} & {\color{darkred} \xmark} & {\color{darkgreen} \cmark} & {\color{darkgreen} \cmark} &  36  & 423 & 417  & -   & 46.00 & 3.64 & $1.75^{-5}$        & 0.57 & 0.86 & 0.65 & 0.63 & 2271652 & 8.24 & 2   \\

{\color{darkgreen} \cmark} & {\color{darkgreen} \cmark} & {\color{darkred} \xmark} & {\color{darkred} \xmark} & {\color{darkgreen} \cmark} & {\color{darkgreen} \cmark} & 70  & 221 & 620  & -   & 98    & 5.15 &  $10^{-2}$            & 0.90 & 0.87 & 0.84 & 0.75 & 2038202 & 8.26 & 1   \\

{\color{darkgreen} \cmark} & {\color{darkgreen} \cmark} & {\color{darkgreen} \cmark} & {\color{darkred} \xmark} & {\color{darkgreen} \cmark} & {\color{darkgreen} \cmark} &121 & 355 & 1955 & 49 & 6.48 & 4.51 & $3.32^{-3}$ & 0.87 & 0.90 & 0.77 & 0.84 & 3584370 & 8.33 & 1\\
{\color{darkred} \xmark} & {\color{darkred} \xmark} & {\color{darkred} \xmark} & {\color{darkred} \xmark} & {\color{darkgreen} \cmark} & {\color{darkgreen} \cmark} &  44  & 622 & 633  & -   & 96.00 & 4.29 & $1.49^{-5}$        & 0.92 & 0.90 & 0.53 & 0.63 & 2541217 & 6.62 & 1 \\

{\color{darkgreen} \cmark} & {\color{darkgreen} \cmark} & {\color{darkgreen} \cmark} & {\color{darkgreen} \cmark} & {\color{darkred} \xmark} & {\color{darkgreen} \cmark} & 134 & -   & 1489 & 36  & 9.51  & 2.44 & $3.87^{-3}$       & 0.50 & 0.97 & -    & 0.57 & 3575787 & 6.93 & 9    \\
{\color{darkgreen} \cmark} & {\color{darkgreen} \cmark} & {\color{darkgreen} \cmark} & {\color{darkgreen} \cmark} & {\color{darkgreen} \cmark} & {\color{darkred} \xmark} &  41  & 356 & -    & 44  & 4.70  & 5.94 & $2.16^{-5}$        & 0.66 & 0.94 & 0.97 & -    & 1700229 & 2.64 & 1  \\
\bottomrule
\end{tabular}
}
\caption{HPs used for evaluation on the ARRAU-AA test set.}
\label{tab:hp2}
\end{table*}

\section{Hyperparameter details}

Tables \ref{tab:hp1} and \ref{tab:hp2} report the tuned HPs for resolution of the shell noun \textit{reason} and resolution of abstract anaphors in ARRAU-AA for different model variants. Below is the list of all tunable HPs.

\begin{itemize}
\item the dimensionality of the hidden states in the bi-LSTM, $h_{LSTM}$ 
\item the first feed-forward layer size, $h_{ffl1}$
\item the second feed-forward layer size, $h_{ffl2}$ 
\item the dimensionality of the tag embeddings, $d_{TAG}$
\item gradient clipping value, $g$
\item frequency of words in vocabulary, $f_w$ 
\item regularization coefficient, $r$ 
\item keep probability of outputs of bi-LSTM, $k_{LSTM}$ 
\item keep probability of input, $k_{input}$
\item keep probability of outputs of the first feed-forward layer, $k_{ffl1}$
\item keep probability of second of the first feed-forward layer, $k_{ffl2}$
\end{itemize}

We additionally report the number of trainable parameters (\# param\.), the average epoch training time using one Nvidia GeForce GTX1080 gpu ($t_e$) and the epoch after which the best score is achieved ($e$).

\end{document}